\documentclass[10pt,twocolumn,letterpaper]{article}

\usepackage[iccvfinal]{iccv}      
\usepackage{algorithm}
\usepackage{algorithmic}
\usepackage{amsfonts}
\usepackage{amssymb}
\usepackage{amsmath}
\usepackage{graphicx}
\usepackage{tabularx}
\usepackage{booktabs}
\usepackage{multirow}
\usepackage{makecell}
\usepackage{booktabs}
\usepackage{array}
\definecolor{iccvblue}{rgb}{0.21,0.49,0.74}
\usepackage[pagebackref,breaklinks,colorlinks,allcolors=iccvblue]{hyperref}

\def\paperID{15407} 
\def\confName{ICCV}
\def\confYear{2025}

\title{Similarity Memory Prior is All You Need for Medical Image Segmentation}

\author{Hao Tang\textsuperscript{1}, Zhiqing Guo\textsuperscript{1,2,\thanks{Corresponding author}}, Liejun Wang\textsuperscript{1,3,*}, Chao Liu\textsuperscript{1}\\
$^{1}$School of Computer Science and Technology, Xinjiang University\\
$^{2}$Xinjiang Multimodal Intelligent Processing and Information Security \\Engineering Technology Research Center\\
$^{3}$Silk Road Multilingual Cognitive Computing International Cooperation \\Joint Laboratory, Xinjiang University\\
{\tt\small \{107552201330@stu., guozhiqing@, wljxju@, 107556522208@stu.\}xju.edu.cn}
}

\begin{document}
\maketitle
\begin{abstract}
In recent years, it has been found that “grandmother cells” in the primary visual cortex (V1) of macaques can directly recognize visual input with complex shapes. This inspires us to examine the value of these cells in promoting the research of medical image segmentation. In this paper, we design a Similarity Memory Prior Network (Sim-MPNet) for medical image segmentation. Specifically, we propose a Dynamic Memory Weights-Loss Attention (DMW-LA), which matches and remembers the category features of specific lesions or organs in medical images through the similarity memory prior in the prototype memory bank, thus helping the network to learn subtle texture changes between categories. DMW-LA also dynamically updates the similarity memory prior in reverse through Weight-Loss Dynamic (W-LD) update strategy, effectively assisting the network directly extract category features. In addition, we propose the Double-Similarity Global Internal Enhancement Module (DS-GIM) to deeply explore the internal differences in the feature distribution of input data through cosine similarity and euclidean distance. Extensive experiments on four public datasets show that Sim-MPNet has better segmentation performance than other state-of-the-art methods. Our code is available on https://github.com/vpsg-research/Sim-MPNet.
\end{abstract}    
\section{Introduction}
\label{sec:intro}
Medical image segmentation, as a basic and key technology, is increasingly highlighting its importance in disease diagnosis, treatment planning, and curative effect evaluation \cite{a1,1111, 2222}. With the rapid development of artificial intelligence and computer vision, constructing automatic and accurate medical image segmentation methods has become a research hotspot \cite{a339, a340}. These methods aim to accurately identify and segment the anatomical structures or pathological features of interest from medical images, to provide clinicians with more intuitive and quantitative medical information \cite{a2,a3}.

\begin{figure}[t]
\centerline{\includegraphics[width=0.5\textwidth]{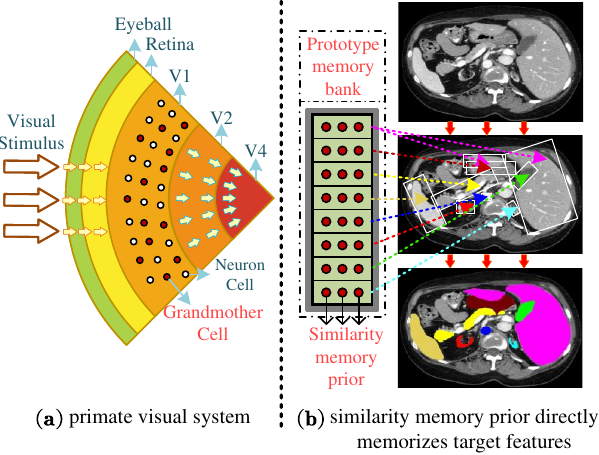}}
\caption{\textbf{Motivation.} (a) Schematic diagram of the primate visual system. (b) We use similarity memory priors stored in the prototype memory bank to imitate “grandmother cells” in V1, and utilize these features to directly match various organs in medical images.}
\label{figure1}
\end{figure}
\begin{figure*}[t]
\centerline{\includegraphics[width=1.0\textwidth]{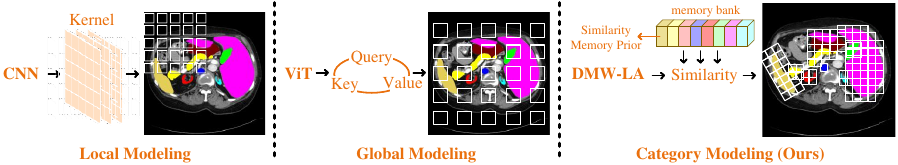}}
\caption{\textbf{Compared with the existing works.} CNN passively responds to local texture through convolution kernel. ViT only focuses on high-frequency textures through self-attention. DMW-LA uses similarity memory priors to directly identify and extract category features of organs.}
\label{figure1-1}
\end{figure*}
In existing medical image segmentation works, researchers attempt to use Convolutional Neural Network (CNN) or Vision Transformer (ViT) to enable networks to generate images with precise labels \cite{i2, i3, i21, ax1, a333}. However, although the local receptive field design of CNNs conform to the translational invariance of images, its inductive bias is essentially a passive response to local textures (edges and corners), rather than actively modeling semantic objects (lesions and organs). ViTs require massive data to learn semantic associations of position encoding, while medical image annotation is scarce, leading to model degradation into “pseudo-global modeling” that focuses on high-frequency texture rather than anatomical structure. In addition, CNNs and ViTs belong to data-driven “implicit learning” and lack prior knowledge to quickly identify key features. Recently, researchers have used two-photon calcium imaging technology to discover the presence of curvature detecting cells in the V1 \cite{a7, a8}. These curvature detecting cells can be regarded as a type of “grandmother cells” with specific activation conditions, as shown in \Cref{figure1}(a). The “grandmother cells” have strong recognition abilities and only respond to specific input stimuli \cite{a400, a401, a402}. If the design of medical segmentation networks is not limited to traditional calculation mechanisms and directly focuses on lesions or organs (category features), it may be a promising design paradigm.


Inspired by this, we try to construct the prototype memory bank with different similarity memory priors to simulate the biological mechanism of “grandmother cells”, as shown in \Cref{figure1}(b). Specifically, similarity memory priors are matched with organ features in medical images through similarity, and a multi-dimensional mapping relationship based on similarity is constructed to extract rich category features in medical images. Meanwhile, this mapping relationship is strengthened by updating the prototype memory bank. In this process, the prototype memory bank is equivalent to a dynamic prior knowledge repository. With the continuous input of new organ features, the similarity memory priors will be continuously optimized to match organ category features more accurately. 

In this paper, we design the Dynamic Memory Weights–Loss Attention (DMW-LA) to directly extract rich category features from medical images. DMW-LA accurately matches the similarity memory priors with the category information in medical images through similarity. This process enables the network to directly capture the abstract category information in medical images, thus significantly enhancing the network's segmentation ability. In addition, to ensure that the similarity memory priors in the prototype memory bank can accurately identify lesions or organs, we design a novel Weight-Loss Dynamic (W-LD) updating strategy that dynamically updates the prototype memory bank through feature weights and training losses to maintain the effectiveness of similarity memory priors. To deeply explore the hidden patterns and structures in features, we design a Double-Similarity Global Internal Enhancement Module (DS-GIM) to help the network better understand and distinguish subtle differences between different lesions or organs. Finally, we distill the above insights, and construct a Similarity Memory Prior Network (Sim-MPNet) with dual encoder structure by introducing MaxViT \cite{a341} to strengthen the learning of global context. In summary, the main contributions of this paper can be summarized as follows:
\begin{itemize}
\item We design a novel Similarity Memory Prior Network (Sim-MPNet), which can fully integrate and utilize category features and global context in medical images, thus improving the accuracy in segmentation tasks. 
\item We design Dynamic Memory Weights–Loss Attention (DMW-LA). DMW-LA utilizes prototype memory bank to recognize and extract category information, and dynamically adjusts the semantic cluster features through Weight-Loss Dynamic (W-LD) update strategy. This method is a novel feature extraction paradigm that provides a new perspective for the design of medical image segmentation networks.
\item We design a Double-Similarity Global Internal Enhancement Module (DS-GIM) to help the network better understand the content of medical images. DS-GIM utilizes cosine similarity and euclidean distance to finely distinguish and capture subtle differences between lesions or organs, thus enhancing the network representation ability.
\end{itemize}

\section{Related Work}
\label{sec:formatting}
\textbf{Medical Image Segmentation.} In medical image segmentation, a common method is to segment the image with a full convolutional neural network (FCNN) \cite{a344, a345}. Especially after the landmark network U-Net \cite{i2} came out, researchers successively launched a series of similar U-shaped Convolutional Neural Network (CNN), such as U-Net++ \cite{i3}, U-Net3+ \cite{i4} and nnU-Net \cite{a342}. However, CNNs tend to explore the correlation between local elements, which results in poor performance in segmenting large lesions or organs. Therefore, some researchers use Vision Transformer (ViT) \cite{a12} for segmentation, such as MEDT \cite{i9}, DAE-former \cite{i11} and DTMFormer \cite{i10}. The self-attention mechanism in ViT focuses on information interaction from a global perspective when processing image information, which lacks exploration of the relationships between local elements. Therefore, some scholars combine CNNs with ViTs to improve the local and global modeling ability of networks, such as TransUNet \cite{ax1}, MixTrans \cite{i12}, UCTransNet \cite{i13} and Transfuse \cite{a343}. However, this strategy of integrating local details and global context may be difficult to capture useful category features directly, thus limiting the segmentation performance, as shown in \Cref{figure1-1}. Different from the traditional computing paradigm, our method directly extracts and remembers the representative category information from images, which realizes the accurate segmentation of the target contour.
\\\textbf{Prototype Memory Mechanism.} The memory bank can capture the cross-time features in network training and has been proven effective in other fields. Santoro et al. \cite{i22} used a memory bank to store new training data for few-shot learning, while Yang and Chan \cite{i23} employed it to memorize long-term object information for video tracking. Additionally, Caron et al. \cite{a21} utilized memory queues to accumulate negative samples for unsupervised learning, and Li et al. \cite{i1} proposed Memory Concept Attention (MOCA) to improve the quality of image generation. However, medical images contain various lesions and organs structures, and these structures exhibit significant differences in different individuals and different disease states. For example, the shape, size, position and surrounding tissues of tumors in medical images are ever-changing. The existing prototype memory mechanism lacks sufficient flexibility to update the representation and memory of different structural features in a timely manner. In our work, a novel dynamic memory prior mechanism is designed, which is called DMW-LA. DMW-LA utilizes prototype memory bank to effectively identify the category features of lesions or organs, and dynamically updates the memory bank through the W-LD update strategy. In addition, DS-GIM is also designed to enhance the difference of feature distribution between different lesions or organs.

\begin{figure}[t]
\centerline{\includegraphics[width=0.5\textwidth]{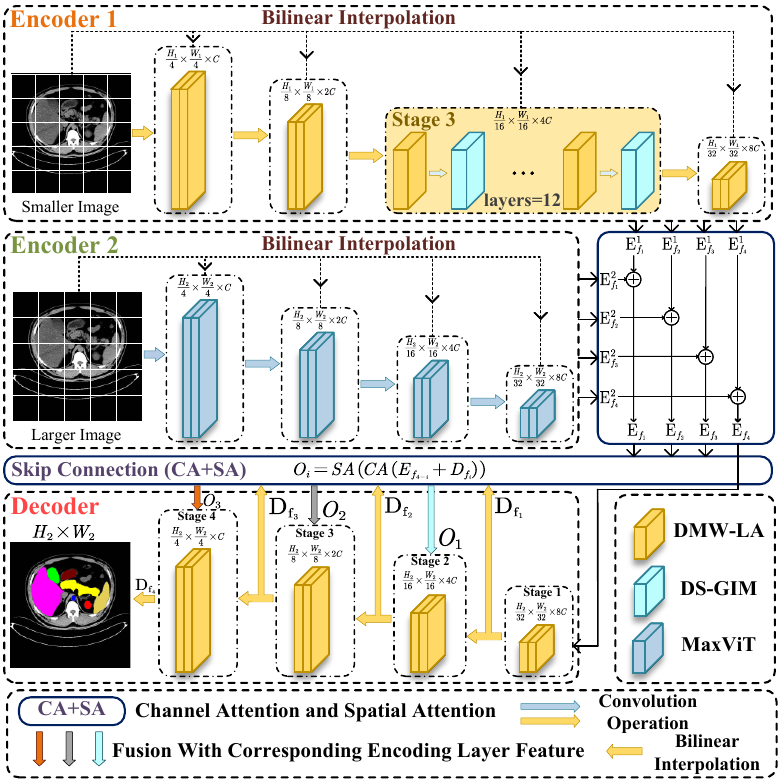}}
\caption{\textbf{Overview of Sim-MPNet}. The network consists of two encoders, one decoder and one skip connection, and it efficiently models intra-class and global dependencies of input features through the dual encoder structure.}
\label{figure2}
\end{figure}

\begin{figure*}[!t]
\centerline{\includegraphics[width=1.0\textwidth]{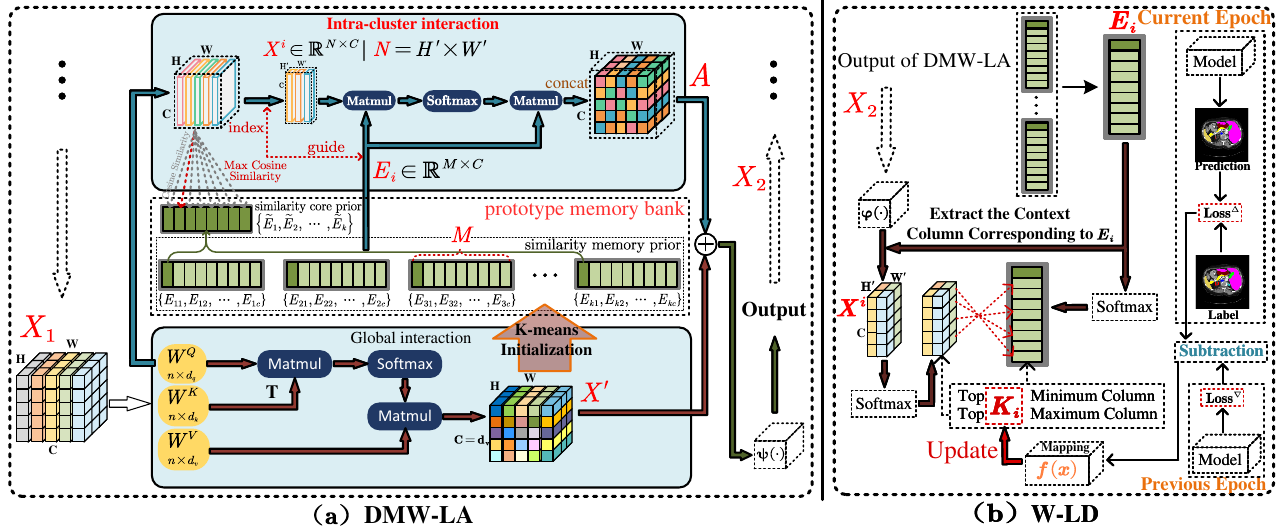}}
\caption{\textbf{Overview of DMW-LA and W-LD update strategy}. (a) DMW-LA consists of intra-cluster interaction and global interaction. (b) The W-LD update strategy adopts weight and loss dual guidance to update the prototype memory bank.}
\label{figure3}
\end{figure*}

\section{Methodology}
\subsection{Overview}
\Cref{figure2} shows the overall structure of the Sim-MPNet. Among them, the encoder 1 consists of four stages DMW-LA. We alternately stack DMW-LA and DS-GIM in the third stage. The encoder 2 consists of a four-stage visual backbone MaxViT \cite{a341}. Each encoder uses overlapping convolution operations to achieve down-sampling. The input image will be re-fused with the stage features of each encoder through bilinear interpolation to reduce information loss caused by resolution reduction. After the encoder completes feature extraction, the network fuses the corresponding stage features of the two encoders, as shown in \Cref{eq21}. 
\begin{equation}
\begin{aligned}
{E}_{{f}_{i}} = {E}_{{f}_{i}}^{1} + {E}_{{f}_{i}}^{2}
\end{aligned}
\label{eq21}
\end{equation}
Where $i=\{1,2,3,4\}$, indicating the encoder stage. ${E}_{{f}_{i}}^{1}$ and ${E}_{{f}_{i}}^{2}$ represent the features of encoder 1 and encoder 2 in stage $i$ respectively. Then, the ${E}_{{f}_{1}}$, ${E}_{{f}_{2}}$ and ${E}_{{f}_{3}}$ is passed into the skip connection, ${E}_{{f}_{4}}$ as the input of the decoder. In skip connection, we connect channel attention (CA) and spatial attention (SA) \cite{i24} in series to fuse the corresponding features of the encoder and decoder, as shown in \Cref{eq22}. 
\begin{equation}
\begin{aligned}
{O}_{i}=SA(CA({E}_{{f}_{4-i}}+{D}_{{f}_{i}}))
\end{aligned}
\label{eq22}
\end{equation}
Where $i=\{1,2,3\}$, ${D}_{{f}_{i}}$ represents the output of decoder in stage $i$, and ${O}_{i}$ represents the output of skip connection. Then, ${O}_{i}+{D}_{{f}_{i}}$ is used as the input of the next stage. The decoder uses DMW-LA to extract deep semantic information and bilinear interpolation to restore the image to the same resolution as the input image to achieve pixel-level segmentation. Next, we will introduce each module in detail.

\subsection{Dynamic Memory Weight-Loss Attention}
In this paper, we design Dynamic Memory Weight–Loss Attention (DMW-LA) to extract key category features from medical images, as shown in \Cref{figure3}(a). The intra-cluster interaction is responsible for obtaining internal dependencies of category features, while global interaction is responsible for initializing prototype memory bank and modeling global dependencies. Assuming input image ${X}_{1}\in {\mathbb{R}}^{H\times W\times C}$, where $H$, $W$, and $C$ refer to the height, width, and channel dimensions of the image, respectively. First, ${X}_{1}$ is linearly transformed with weights ${W}^{q}$, ${W}^{k}$ and ${W}^{v}$ to generate Query, Key, and Value for global interaction. At the beginning of network training, we perform K-means on the result ${X}^{\prime}$ of global interaction to fill the memory bank. Here, we call the generated semantic clusters as similarity memory prior, and the central point of each cluster is called the similarity core prior. Therefore, the similarity memory prior can be represented as the $\left \{ \left \{ {E}_{11},{E}_{12},\cdots ,{E}_{1c}\right \},\cdots ,\left \{ {E}_{k1}, {E}_{k2},\cdots ,{E}_{kc}\right \}\right \}$, and the similarity core prior can be represented as the $\left \{ \tilde{{E}_{1}},\tilde{{E}_{2}},\tilde{{E}_{3}},\cdots ,\tilde{{E}_{k}}\right \}$. After the initialization of the memory bank is completed, we can supplement prior knowledge for network training. Here, we first calculate the cosine similarity between the Query and the similarity core prior to match the corresponding similarity memory prior. Assuming there is $\left \{{X}^{i}\in {\mathbb{R}}^{N\times C},N\leqslant H\times W\right \}$ in the Query that matches the similarity memory prior $\left \{ {E}_{i}={E}_{i1},{E}_{i2},\cdots ,{E}_{ic}\mid {E}_{i}^{T}\in {\mathbb{R}}^{C\times M}\right \}$, then ${E}_{i}$ will perform self-attention with ${X}^{i}$, as shown in \Cref{eq23}. 
\begin{equation}
\begin{aligned}
{A}_{i}=SM({X}^{i}\cdot {E}_{i}^{T})\cdot {E}_{i}
\end{aligned}
\label{eq23}
\end{equation}
Where $i=\{1,2,3,...\}$, indicating the number of segmented categories, and $SM$ represents the softmax activation function. This step simulates the process of the “grandmother cell” recognizing specific visual stimuli, so the ${E}_{i}$ plays both the role of Key and Value. Then, we concatenate ${A}_{i}$ to obtain semantic category features $A$. Finally, we aggregate semantic category features and global features, and then map to the original feature space through the convolution operation $\psi (\cdotp)$, as shown in \Cref{eq24}.
\begin{equation}
\begin{aligned}
{X}_{2}=\psi (A+{X}^{\prime})
\end{aligned}
\label{eq24}
\end{equation}



After the initialization of the memory bank is completed, it still needs to continuously update it to ensure that it can adapt to complex medical images. At present, the common method is to use a random update strategy, that is, randomly insert a portion of the input features into the corresponding semantic clusters in the memory. This method is not suitable for complex and ever-changing medical images and may introduce interference information. Therefore, we propose a Weight-Loss Dynamic (W-LD) update strategy to dynamically update the memory bank, as shown in \Cref{figure3}(b). Specifically, assume that ${X}^{i}\in {\mathbb{R}}^{N\times C}$ in the output ${X}_{2}\in {\mathbb{R}}^{H\times W \times C}$ of the DMW-LA matches the similarity memory prior ${E}_{i}\in {\mathbb{R}}^{M\times C}$. Then use the Softmax to calculate the weights of ${X}^{i}$ and ${E}_{i}$ separately, and sort the columns of ${X}^{i}$ and ${E}_{i}$ respectively according to the weight size. Note that the columns with large weight in ${X}^{i}$ contain more semantic information, while the columns with small weight in ${E}_{i}$ contain less semantic information. Therefore, we replace the $K$ columns in ${E}_{i}$ with the smallest weight with the $K$ columns in ${X}^{i}$ with the largest weight. In addition, we evaluate the quality of the current prototype memory bank by calculating the difference between the current epoch loss (${Loss}^{\varDelta}$) and the previous epoch loss (${Loss}^{\nabla}$). Specifically, when the value of ${Loss}^{\nabla}-{Loss}^{\varDelta}$ is greater than 0, it shows that the prototype memory bank has a positive impact on network training, leading to the reduction of the current epoch loss. Therefore, the $K$ should be reduced to preserve these valuable features. Conversely, when the value of ${Loss}^{\nabla}-{Loss}^{\varDelta}$ is less than 0, the $K$ should be increased to expand the range of updates and eliminate useless features as much as possible. In this way, the prototype memory bank can accumulate high-value semantic information to the maximum extent in each training epoch. In addition, a linear function is designed to link $K$ and ${Loss}^{\nabla}- {Loss}^{\varDelta}$, as shown in \Cref{eq1}.

\begin{figure}[t]
\centerline{\includegraphics[width=0.49\textwidth]{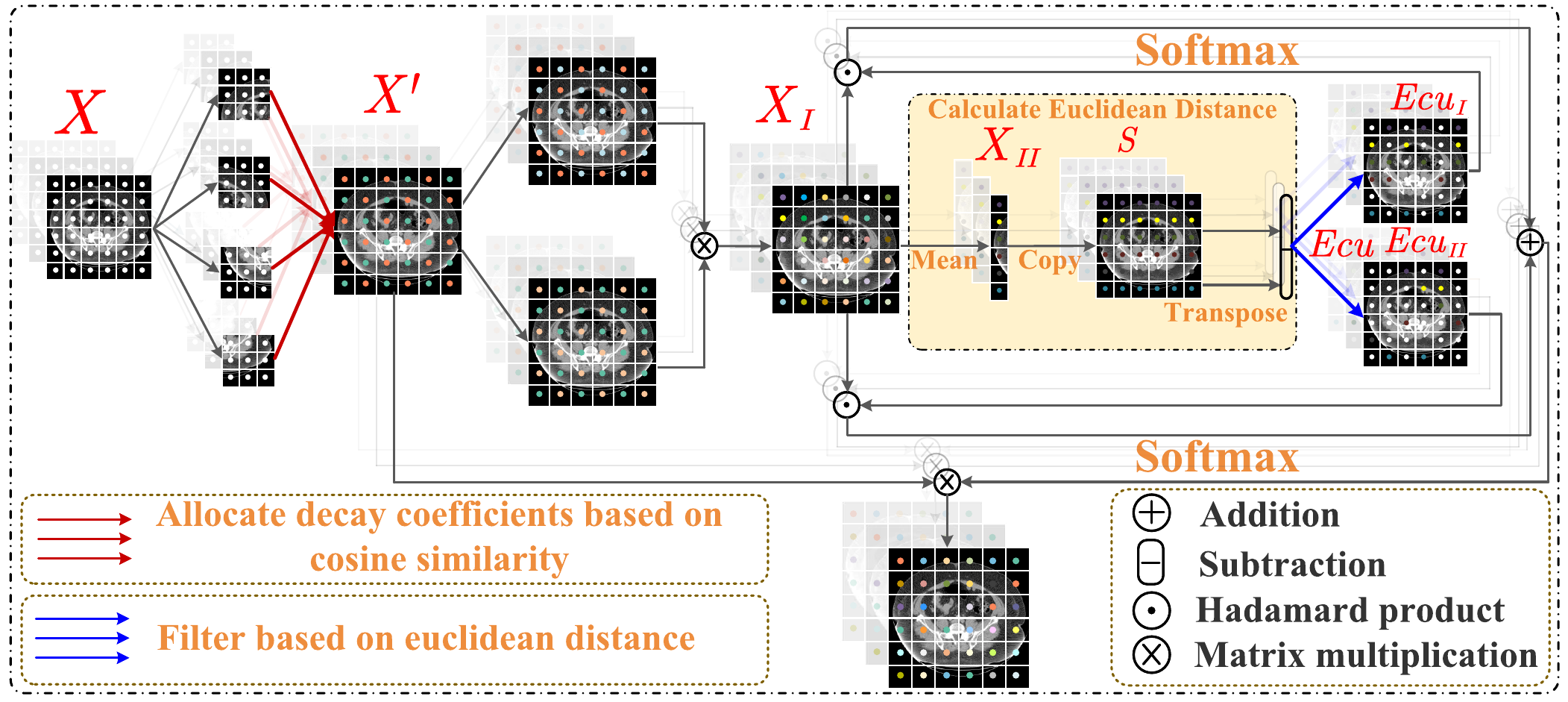}}
\caption{\textbf{Overview of DS-GIM}. DS-GIM uses cosine similarity and euclidean distance to comprehensively evaluate and highlight the differences between spatial element structures.}
\label{figure5}
\end{figure}

\begin{equation}
\begin{aligned}
K=(-\alpha \cdot ({Loss}^{\nabla}-{Loss}^{\varDelta})+\beta )\cdot \theta 
\end{aligned}
\label{eq1}
\end{equation}
In the above formula, $\alpha$ and $\beta$ are decay factors, which are set to 0.5, and $\theta$ = $M$. In addition, we set $\frac{M}{4}\leq  K\leq  \frac{3M}{4}$ to avoid $K$ falling into extreme situations and ensure stable training, see appendix A. In this way, we can not only minimize the loss of information but also ensure the semantic clarity and representation ability of the similarity memory prior. 

\subsection{Double-Similarity Global Internal Enhancement Module}
The structure of the Double Similarity Global Internal Enhancement Module (DS-GIM) is shown in \Cref{figure5}. Assuming the input size is $X\in {\mathbb{R}}^{H\times W\times C}$, we first divide $X$ into $\left \{{X}_{1},{X}_{2},\cdots ,{X}_{\frac{H\times W}{{p}^{2}}}\mid {X}_{i}\in {\mathbb{R}}^{P\times P\times C}\right \}$ with the windows of $P\times P$, and calculate the internal cosine similarity of these windows. Weight different decay coefficients for each window based on similarity (from large to small). The purpose is to use the decay coefficient to increase the difference between windows. The formula for calculating the decay coefficient is as follows. 
\begin{equation}
\begin{aligned}
\gamma  = \exp\left( -\left( 0.25 - 2^{-2.5 - \frac{5 \cdot \mathbf{n}}{\text{$l$}}} \right) \right) 
\end{aligned}
\label{eq2}
\end{equation}
In the above formula, $l$ represents the total number of layers in the current stage, $n=\{0, 1, 2, \cdots, \text{$l$}-1\}$. Next, concatenate these windows to form a new feature representation ${X}^{'}$. Map ${X}^{'}$ to a new feature space for feature calculation to obtain ${X}_{I}\in {\mathbb{R}}^{HW\times HW}$. Then average the columns of ${X}_{I}$ to get ${X}_{II}\in {\mathbb{R}}^{HW\times 1}$, and then copy ${X}_{II}$ to get a new matrix $S \in {\mathbb{R}}^{HW\times HW}$. At this time, the elements in $S$ can be regarded as a set of spatial location elements, and we can calculate the euclidean distance between elements using the formula $Euc=\left | S-{S}^{T}\right |$ to measure their proximity in the feature space. Based on size, we can divide the internal elements of $Euc$ into ${Ecu}_{I}$ and ${Ecu}_{II}$. Next, these two parts are calculated by the Softmax function and are hadamard products with ${X}_{I}$ respectively. Finally, splice these two parts together to form a new matrix and multiply it with ${X}^{'}$, as shown in \Cref{eq25}. 
\begin{equation}
\begin{aligned}
{O}_{D}={X}^{\prime}\cdot (SM({Ecu}_{I})*{X}_{I}+SM({Ecu}_{II})*{X}_{I}) 
\end{aligned}
\label{eq25}
\end{equation}
In the above equation, $*$ represents hadamard product, and $SM$ represents softmax activation function. In this way, distance information between spatial elements will be embedded into feature encoding to enhance the model's understanding of the internal structure of input data.



\begin{table*}[ht]
\small 
\renewcommand{\arraystretch}{1.05} 
\setlength{\tabcolsep}{1.2mm} 
\centering
\begin{tabular}{c|c|ccccc|cccc|cc}
\specialrule{0.04em}{0pt}{0pt}
\midrule
\multirow{2}{*}{Type} & \multirow{2}{*}{Methods} & \multicolumn{5}{c|}{\textbf{ACDC}} & \multicolumn{4}{c|}{\textbf{SegPC-2021}} & \multicolumn{2}{c}{\textbf{ISIC-2018}} \\
\cline{3-13}
& & DSC↑ & HD95↓ & RV & Myo & LV & DSC↑ & HD95↓ & Cytoplasm & Nucleus & DSC↑ & HD95↓ \\
\midrule
\specialrule{0.04em}{0pt}{0pt}
\multirow{7}{*}{CNN}& U-Net \cite{i2} & 90.14 & 1.48 & 89.61 & 86.58 & 94.23 & 80.54 & 35.89 & 80.77 & 80.30 & 89.49 & 4.66\\
 &Att-UNet \cite{a18} & 90.48 & 1.24 &89.15 &87.29 &95.01 & 80.40 & 36.04 & 79.56 & 81.37 & 89.58 & 5.11 \\
 &U-Net++ \cite{i3} & 89.54 & 1.25 &87.97 &86.62 &94.05 & 80.07 & 33.18 & 79.40 & 80.12 & 89.19 & 4.88 \\
 &EMCAD \cite{a20} & 91.43 & 1.13 &90.65 &88.78 &94.86 & 80.38 & 34.84 & 79.49 & 81.27 & 90.03 & 4.33 \\
 &MADGNet \cite{re3} & 91.13 & 1.17 &90.01 &88.30 &95.07 & 80.02 & 33.21 & 79.55 & 80.48 & 89.77 & 4.30 \\
 &SelfReg-UNet \cite{re4} & 91.43 & 1.19 &88.92 &\textbf{89.49} &\textbf{95.88} & 77.60 & 35.65 & 77.44 & 77.75 & 88.99 & 5.74 \\
 & DconnNet \cite{re5} & 89.96 & 1.80 &89.16 &86.12 &94.58 & 79.84 & 33.07 & 79.00 & 80.68 & 90.01 & 4.10 \\
\midrule
\multirow{4}{*}{ViT} & SwinUNet \cite{i21} & 89.49& 2.03   &88.01 &86.03 &94.45 & 78.22 & 35.24 & 77.33 & 79.11 &    87.32 & 5.47 \\
 &TransDeepLab \cite{i19} & 89.22 & 2.07 &87.48 &85.74 &94.44 & 78.81 & 33.02 & 77.38 & 80.24 & 90.11 & 4.07 \\
 &DAE-former \cite{i11} & 89.78 & 1.91 &89.91 &84.38 &95.04 & 78.52 & 34.86 & 77.10 & 79.93 & 89.85 & 4.22 \\
 & SelfReg-SwinUNet \cite{re4} & 91.49 & 1.97 &89.49 &89.27 &95.70 & 76.40 & 32.45 & 74.70 & 78.09 & 89.16 & 4.90 \\
 \midrule
 \multirow{6}{*}{\makecell{CNN\&\\ViT}}&TransUNet \cite{ax1} &89.37 & 1.48 &89.20 &85.23 &93.67& 76.69 & 33.95 & 78.69 & 80.68 & 89.92 & 4.34 \\
 &MixTrans \cite{i12} & 90.98 & 2.28 &89.52 &88.60 &94.82& 80.72 & 32.62 & 79.78 & 81.66 & 86.46 & 7.34 \\
 &TransClawUNet \cite{i17} & 91.29 & 1.10 &91.31 &87.93 &94.64& 79.77 & 34.45 & 79.04 & 80.50 & 89.62 & 4.81 \\
 &HiFormer \cite{i18} & 90.12 & 2.15 &91.06 &84.54 &94.77 & 79.54  & 32.08 & 77.73 & 81.35 & 89.79  & 4.58 \\
 &TransCASCADE \cite{i20} & 91.48 & 1.12 &91.37 &88.32 &94.76& 80.36 &  32.87 & 79.18 &  81.53 & 89.76 &  4.42 \\
 &PAG-TransYnet \cite{a19} & 91.33 & 1.15 &90.80 &88.28 &94.92& 79.41 & 33.88 & 78.67 & 80.16 & 89.65 & 4.53 \\
\midrule
 {\makecell{Memory \\ Mechanism}}&Sim-MPNet (\textbf{Ours}) & \textbf{92.18} & \textbf{1.08} &\textbf{92.06} & 89.19 &95.29& \textbf{83.12} & \textbf{29.66} & \textbf{82.00} &  \textbf{84.24} & \textbf{90.85} & \textbf{3.55} \\
\bottomrule
\specialrule{0.04em}{0pt}{0pt}
\end{tabular}
\caption{\label{table1}\textbf{Performance on ACDC, SegPC-2021 and ISIC-2018.} We report the DSC and HD95 scores of different networks on the ACDC, SegPC-2021 and ISIC-2018. We adopt publicly available SOTA methods for implementation, and all results are obtained under the same dataset partitioning conditions. DSC are reported for individual organs or lesions. ↑(↓) denotes the higher (lower) the better.}
\end{table*}
\begin{table*}[ht]
\small 
\renewcommand{\arraystretch}{1.02} 
\setlength{\tabcolsep}{1.0mm} 
\centering
\begin{tabularx}{\linewidth}{c|c|c|c|cc|cccccccc|c}
\specialrule{0.04em}{0pt}{0pt}
\toprule
Type & Methods & P(M) & F(G) & DSC↑ & HD95↓ & Aorta & GB & KL & KR & Liver & PC & SP & SM & ${T}_{i}$ \\
\midrule
\specialrule{0.04em}{0pt}{0pt}
\multirow{7}{*}{CNN}&U-Net \cite{i2} & 31.04 & 50.14 & 76.85 & 39.70 & 89.07 & 69.72 & 77.77 & 68.60 & 93.43 & 53.98 & 86.67 & 75.58 & 729\\
&Att-UNet \cite{a18} & 34.88 & 34.88 & 77.77 & 36.02 & 89.55 & 68.88 & 77.98 & 71.11 & 93.57 & 58.04 & 87.30 & 75.75 & 761\\
&U-Net++ \cite{i3} & 36.63 & 107.01 & 78.98 & 30.65 & 87.45 & 65.86 & 81.75 & 74.27 & 93.85 & 64.16 & 87.82 & 76.65 & 1139\\
&EMCAD \cite{a20} & 26.76 & 4.29 & 83.63 & 15.68 & 88.18 & 68.87 & 88.08 & 84.10 & 95.26 & 68.51 & \textbf{92.17} & 83.92 & 752\\
&MADGNet \cite{re3} & 32.16 & 7.71 & 80.98 & 27.46 & 88.01 & 65.12 & 82.81 & 79.09 & 94.92 & 67.23 & 90.15 & 80.50 & 1152\\
&SelfReg-UNet \cite{re4} & 17.26 & 30.71 & 80.34 & 26.98 & 88.74 & 71.78 & 85.32 & 80.71 & 93.80 & 62.21 & 84.78 & 75.39 & 625\\
&DconnNet \cite{re5} & 36.40 & 4.80 & 81.56 & 19.01 & 87.48 & 68.56 & 83.70 & 81.04 & 94.64 & 65.24 & 90.64 & 81.15 & 849\\
\midrule
\multirow{4}{*}{ViT}&SwinUNet \cite{i21} & 27.17 & 5.92 & 79.13 & 21.55 & 85.47 & 66.53 & 83.28 & 79.61 & 94.29 & 56.58 & 90.66 & 76.60 & 625\\
&TransDeepLab \cite{i19} & 28.61 & 17.08 & 80.16 & 21.25 & 86.04 & 69.16 & 84.08 & 79.88 & 93.53 & 61.19 & 89.00 & 78.40& 830\\
&DAE-Former \cite{i11} & 48.07 & 26.07 & 82.43 & 17.46 & 88.96 & 72.30 & 86.08 & 80.88 & 94.98 & 65.12 & 91.94 & 79.19& 1187\\
&SelfReg-SwinUNet \cite{re4} & 27.15 & 5.92 & 80.54 & 20.10 & 86.07 & 69.65 & 85.12 & 82.58 & 94.18 & 61.09 & 87.42 & 78.22& 1014\\
\midrule
\multirow{6}{*}{\makecell{CNN\&\\ViT}}&TransUNet \cite{ax1} & 105.32 & 24.66 & 77.48 & 31.69 & 87.23 & 63.13 & 81.87 & 77.02 & 94.08 & 55.86 & 85.08 & 75.62& 794\\
&MixTrans \cite{i12}& 79.07 & 44.75 & 78.59 & 26.59 & 87.92 & 64.99 & 81.47 & 77.29 & 93.06 & 59.46 & 87.75 & 76.81& 958\\
&TransClawUNet \cite{i17} & 113.02 & 38.08 & 78.09 & 26.38 & 85.87 & 61.38 & 84.83 & 79.36 & 94.28 & 57.65 & 87.74 & 73.55& 1253\\
&HiFormer \cite{i18} & 29.52 & 11.88 & 80.69 & 19.14 & 87.03 & 68.61 & 84.23 & 78.37 & 94.07 & 60.77 & 90.44 & 82.03& 931\\
&TransCASCADE \cite{i20}& 123.47 & 56.93 & 82.68 & 17.34 & 86.63 & 68.48 & 87.66 & 84.56 & 94.43 & 65.33 & 90.79 & 83.52& 1472\\
&PAG-TransYnet \cite{a19} & 144.22 & 33.64 & 83.43 & 15.82 & \textbf{89.67} & 68.89 & 86.74 & 84.88 & \textbf{95.87} & 68.75 & 92.01 & 80.66 &940\\
\midrule
{\makecell{Memory \\ Mechanism}}& Sim-MPNet (\textbf{Ours}) & 91.30 & 43.74 & \textbf{84.34} & \textbf{14.85} & 87.06 & \textbf{73.49} & \textbf{88.65} & \textbf{85.34} & 95.28 & \textbf{69.25} & 91.36 & \textbf{84.30} &873\\
\bottomrule
\specialrule{0.04em}{0pt}{0pt}
\end{tabularx}
\caption{\label{table2}\textbf{Performance on Synapse.} P and F represent Parameters and FLOPs respectively. FLOPs of all the methods are reported for 224$\times$224 inputs, except Sim-MPNet (dual encoder: 224$\times$224, 256$\times$256). ${T}_{i}$ is the average inference efficiency of a single image in milliseconds. DSC are reported for individual organs. GB, KL, KR, PC, SP and SM represent gallbladder, left kidney, right kidney, pancreas, spleen and stomach respectively.}
\end{table*}
\section{Experiments}
\label{sec:1}
\subsection{Datasets}
We conduct extensive experiments on four publicly available datasets: Automated Cardiac Diagnosis Challenge (ACDC, 4 classes) \cite{a16}, Segmentation of Multiple Myeloma Plasma Cells in Microscopic Images (SegPC-2021, 3 classes) \cite{a14, a346, a347}, International Skin Imaging Collaboration 2018 Challenge (ISIC-2018, 2 classes) \cite{a15}, and MICCAI 2015 Multi-Atlas Abdomen Labeling Challenge (Synapse, 9 classes). (1) ACDC dataset consists of Magnetic Resonance Imaging (MRI) of the left ventricle (LV), right ventricle (RV) and myocardium (Myo). The division of training set and testing set is consistent with TransUNet++ \cite{a348}. (2) SegPC-2021 dataset consists of myeloma plasma cell images. The segmentation target is the cytoplasm and nucleus of the cells. We divide the training set and the testing set by 4:1 ratio. (3) ISIC-2018 dataset consists of dermoscopy images. We also divide the training set and the testing set by 4:1 ratio. (4) Synapse dataset consists of axial abdominal clinical CT images. Each image contains part or all of the eight abdominal organs (aorta, gallbladder, left kidney, right kidney, liver, pancreas, spleen, stomach). We divide the training set and the testing set according to the TransUNet \cite{ax1}.  


\begin{figure*}[!t]
\centerline{\includegraphics[width=1.0\textwidth]{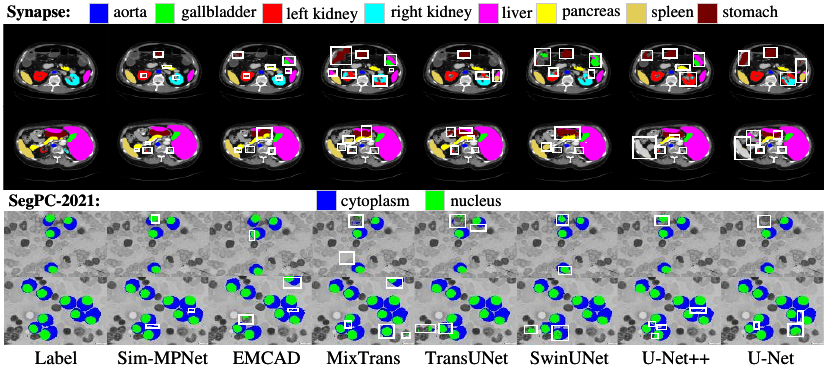}}
\caption{\textbf{Multi-organ and cell segmentation.} Comparison of visualization results between Sim-MPNet and other SOTA methods on Synapse and SegPC-2021. The different segmentation targets are represented by different colors.}
\label{figure6}
\end{figure*}
\subsection{Implementation Details and Evaluation Metrics}
In all experiments, the hyperparameters of DS-GIM (decay coefficient) and DMW-LA (linear function) remained unchanged. For different datasets, just set k in K-means as the number of categories in the dataset. The input resolutions of two encoders are set to 224$\times $224 and 256$\times $256 respectively. In addition, we train the model using an AdamW optimizer with a learning rate of 0.0001, and a weight decay of 1e-4. The model is trained on a single RTX 3090 GPU with a batch size 20 using Dice ($L_{dice}$) and Cross-entropy ($L_{ce}$) losses for 300 epochs. The specific proportion of these two losses is $Loss=0.7\cdot L_{dice}+0.3\cdot L_{ce}$. Finally, all experiments are performed in the Python 3.10.0 and Torch 2.3.0 environment. Only use random rotation and flipping as data augmentation. We choose the Dice score (DSC) and 95\% Hausdorff distance (HD95) as evaluation metrics. The HD95 represents the ${95}^{th}$ percentile distance between the boundaries of the model prediction graph and real labels.

\subsection{Comparison with SOTA Methods}

\textbf{ACDC.} ACDC datasets test the model's ability to explore local details, and ViT-based methods are often difficult to achieve better segmentation results. \Cref{table1} shows that the DSC of SwinUNet, TransDeepLab and DAE-Former does not exceed 90\%. Although the hybrid CNN-ViT methods partially compensate for the shortcomings of ViT, its highest DSC is only 91.48\%. In contrast, Sim-MPNet can directly capture the category features of organs in ACDC by dynamic memory prior mechanism, thus achieving 92.18\% DSC and acquiring better segmentation results.
\\\textbf{SegPC-2021.} \Cref{table1} shows that the Sim-MPNet outperforms existing methods and achieves a new SOTA on SegPC-2021. Because most of the myeloma plasma cells in SegPC-2021 are closely connected, this requires the network to have a strong ability to extract local details. Therefore, we can observe that the CNN-based methods and the hybrid CNN-ViT methods are better than the ViT-based methods in segmentation effect. However, compared with the optimal network MixTrans in the above methods, Sim-MPNet improved DSC by 2.40\% and HD95 by 2.42mm. This proves that Sim-MPNet can effectively handle complex medical image.
\\\textbf{ISIC-2018.} \Cref{table1} shows that the Sim-MPNet also achieved SOTA results on the ISIC-2018 dataset. Since there are many large-scale lesions in the ISIC-2018, it verifies the global modeling ability of the network. Thus, the ViT-based TransDeepLab method has achieved the best segmentation results among the existing methods. However, compared with TransDeepLab method, Sim-MPNet improves DSC by 0.74\% and HD95 by 0.52mm. This proves once again that Sim-MPNet has the advantage of capturing category dependencies.
\\\textbf{Synapse.} To further verify the performance of Sim-MPNet model in different complex scenarios, we introduce more diverse Synapse datasets for evaluation. \Cref{table2} shows the comparison results on the Synapse. Firstly, CNN-based methods have significant advantages in small organ segmentation due to their excellent local modeling ability. For example, EMCAD achieved DSC of 88.08\% and 84.10\% on the left kidney (Kidney (L)) and right kidney (Kidney (R)), respectively. In contrast, Sim-MPNet performed better on these two organs, with DSC of 0.57\% and 1.24\% higher than EMCAD, respectively. However, on the Spleen with a complex contour, Sim-MPNet cannot fully calculate the correlation between local elements, and thus it cannot achieve the optimal segmentation result. Secondly, ViT-based methods have advantages in large-scale organ segmentation due to their global context exploration ability. DAE-Former achieved DSC of 72.30\% and 94.98\% on Gallbladder and Liver, respectively. Sim-MPNet has a DSC of 1.19\% and 0.3\% higher than DAE-Former in these two organs, respectively. Finally, the hybrid CNN-ViT methods combine the advantages of CNN and ViT, enabling the PAG-TransYnet network to explore both local and global features simultaneously, achieving optimal segmentation results in Aorta with the smallest area and Liver with the largest area. Although Sim-MPNet is slightly inferior to PAG-TransYnet in segmentation of the Aorta and Liver, its unique dynamic memory prior mechanism enables it to achieve the best average DSC and HD95 scores on the Synapse. This indicates that Sim-MPNet has excellent robustness in various scenarios. In addition, Sim-MPNet does not have significant advantages in terms of Params and FLOPs, but our method extracts medical image features from a new perspective and achieves excellent segmentation results. This shows the application potential of our method in the field of medical image segmentation and opens up a new path for future research.
\\\textbf{Visual Comparison and Analysis.} \Cref{figure6} shows the visualizations of Sim-MPNet and other networks on Synapse and SegPC-2021. Firstly, the segmentation results of Sim-MPNet are very close to the ground truth in Synapse. Especially when dealing with organs like the Pancreas, which have complex contours and rich details, Sim-MPNet shows better segmentation performance compared to other networks. Secondly, Sim-MPNet still obtains the best segmentation results for images containing multiple cells in SegPC-2021. Although false positives and false negatives may also occur, the incidence of these errors is significantly reduced compared to other networks. In addition, the semantic clustering receptive fields of DMW-LA on Synapse, ACDC and SegPC-2021 see appendix B.

\begin{table}[!t]
\small 
\renewcommand{\arraystretch}{1.4} 
\setlength{\tabcolsep}{1.5mm} 
\centering
\resizebox{\linewidth}{!}{\begin{tabular}{cc|c ccc}
\midrule
\specialrule{0.03em}{0pt}{0pt}
\makecell{Encoder 1}& Encoder 2 & ACDC & SegPC-2021 & ISIC-2018 &  Synapse \\
\midrule
DMW-LA (\textbf{Ours}) & --- & 91.54 & 82.65 & 89.57 & 83.46\\
--- & MaxViT & 91.06 & 81.92 & 89.23 & 83.27\\
MaxViT & MaxViT & 91.33 & 82.07 & 90.04 & 83.59 \\
\midrule
DMW-LA (\textbf{Ours}) & MaxViT & \textbf{92.18} & \textbf{83.12} & \textbf{90.85} & \textbf{84.34}\\
\midrule
\specialrule{0.03em}{0pt}{0pt}
\end{tabular}}
\caption{\label{table3}\textbf{Ablation study of DMW-LA} on four datasets. We use different encoders to verify their effectiveness separately. The input resolution of each encoder is the same as that of the comparative experiment, which is 224$\times$224 (Encoder 1) and 256$\times$256 (Encoder 2) respectively.} 
\end{table}
\subsection{Ablation Study}
\textbf{Effectiveness of DMW-LA.} In this paper, Sim-MPNet adopts dual encoder structure, where Encoder 1 is composed of DMW-LA and Encoder 2 is composed of MaxViT. To evaluate the effectiveness of DMW-LA, we conducted experiments on each encoder. As shown in \Cref{table3}, the obvious performance difference indicates that DMW-LA can help the network obtain rich category features, thereby improving segmentation performance. \\
\textbf{Effectiveness of DS-GIM and W-LD Update Strategy.} We conducted ablation experiments on DS-GIM and W-LD update strategy. It can be seen from \Cref{table4} that compared with the random update strategy, our W-LD update strategy can minimize redundant information and retain useful features. In addition, the remarkable performance improvement in the experiment further confirms the effectiveness of DS-GIM in capturing the internal differences of features.
\begin{table}[!t]
\small 
\renewcommand{\arraystretch}{1.2} 
\setlength{\tabcolsep}{0.9mm} 
\centering
\begin{tabular}{ccccc}
\midrule
\specialrule{0.03em}{0pt}{0pt}
Dataset & ACDC & SegPC-2021 & ISIC-2018 & Synapse \\
\midrule
w/o W-LD & 91.53 & 82.63 & 90.22 & 83.75\\
w/o DS-GIM & 91.67 & 82.58 & 90.47 & 83.64\\
w/o W-LD\&DS-GIM & 91.02 & 82.23 & 90.07 & 83.39\\
\midrule
Sim-MPNet& \textbf{92.18} & \textbf{83.12} & \textbf{90.85} & \textbf{84.34}\\
\midrule
\specialrule{0.03em}{0pt}{0pt}
\end{tabular}
\caption{\label{table4}\textbf{Ablation study of DS-GIM and W-LD update strategy} on four datasets. In w/o W-LD, we use a random update strategy to replace the W-LD update strategy. The figures in the table are DSC.}
\end{table}
\section{Conclusion}
In this paper, we break the limitations of traditional methods and construct a novel medical image segmentation network, namely Sim-MPNet. Sim-MPNet directly extracts and locates the category features of lesions or organs. We develop two core modules: DMW-LA and DS-GIM. DMW-LA obtains category information through prototype memory bank and W-LD update strategy, while DS-GIM enhances the intrinsic differences between elements through a dual similarity mechanism. The comparative experiments show that Sim-MPNet outperforms the previous SOTA methods in segmenting multiple organs, cells, and lesions. We believe that Sim-MPNet brings new perspectives and inspirations to the design of medical image segmentation networks in future.
\section*{Acknowledgements}
This work was supported in part by the Tianshan Talent Training Program under Grant 2022TSYCLJ0036, in part by the National Natural Science Foundation of China under Grant 62472368, Grant 62302427 and Grant 62462060.
\newpage
{
    \small
    \bibliographystyle{ieeenat_fullname}
    \bibliography{main}

\begin{thebibliography}{50}
\providecommand{\natexlab}[1]{#1}
\providecommand{\url}[1]{\texttt{#1}}
\expandafter\ifx\csname urlstyle\endcsname\relax
  \providecommand{\doi}[1]{doi: #1}\else
  \providecommand{\doi}{doi: \begingroup \urlstyle{rm}\Url}\fi

\bibitem[Azad et~al.(2022)Azad, Heidari, Shariatnia, Aghdam, Karimijafarbigloo, Adeli, and Merhof]{i19}
Reza Azad, Moein Heidari, Moein Shariatnia, Ehsan~Khodapanah Aghdam, Sanaz Karimijafarbigloo, Ehsan Adeli, and Dorit Merhof.
\newblock Transdeeplab: Convolution-free transformer-based deeplab v3+ for medical image segmentation.
\newblock In \emph{Int. Workshop PRedict. Intell. Med.}, pages 91--102. Springer, 2022.

\bibitem[Azad et~al.(2023)Azad, Arimond, Aghdam, Kazerouni, and Merhof]{i11}
Reza Azad, Ren{\'e} Arimond, Ehsan~Khodapanah Aghdam, Amirhossein Kazerouni, and Dorit Merhof.
\newblock Dae-former: Dual attention-guided efficient transformer for medical image segmentation.
\newblock In \emph{In Int. Workshop PRedict. Intell. Med.}, pages 83--95. Springer, 2023.

\bibitem[Azad et~al.(2024)Azad, Aghdam, Rauland, Jia, Avval, Bozorgpour, Karimijafarbigloo, Cohen, Adeli, and Merhof]{a333}
Reza Azad, Ehsan~Khodapanah Aghdam, Amelie Rauland, Yiwei Jia, Atlas~Haddadi Avval, Afshin Bozorgpour, Sanaz Karimijafarbigloo, Joseph~Paul Cohen, Ehsan Adeli, and Dorit Merhof.
\newblock Medical image segmentation review: The success of u-net.
\newblock \emph{IEEE Trans. Pattern Anal. Mach. Intell.}, 2024.

\bibitem[Bernard et~al.(2018)Bernard, Lalande, Zotti, Cervenansky, Yang, Heng, Cetin, Lekadir, Camara, Ballester, et~al.]{a16}
Olivier Bernard, Alain Lalande, Clement Zotti, Frederick Cervenansky, Xin Yang, Pheng-Ann Heng, Irem Cetin, Karim Lekadir, Oscar Camara, Miguel Angel~Gonzalez Ballester, et~al.
\newblock Deep learning techniques for automatic mri cardiac multi-structures segmentation and diagnosis: is the problem solved?
\newblock \emph{IEEE Trans. Med. Imaging}, 37\penalty0 (11):\penalty0 2514--2525, 2018.

\bibitem[Bougourzi et~al.(2024)Bougourzi, Dornaika, Taleb-Ahmed, and Truong~Hoang]{a19}
Fares Bougourzi, Fadi Dornaika, Abdelmalik Taleb-Ahmed, and Vinh Truong~Hoang.
\newblock Rethinking attention gated with hybrid dual pyramid transformer-cnn for generalized segmentation in medical imaging.
\newblock In \emph{International Conference on Pattern Recognition}, pages 243--258. Springer, 2024.

\bibitem[Cao et~al.(2022)Cao, Wang, Chen, Jiang, Zhang, Tian, and Wang]{i21}
Hu Cao, Yueyue Wang, Joy Chen, Dongsheng Jiang, Xiaopeng Zhang, Qi Tian, and Manning Wang.
\newblock Swin-unet: Unet-like pure transformer for medical image segmentation.
\newblock In \emph{ECCV}, pages 205--218. Springer, 2022.

\bibitem[Caron et~al.(2020)Caron, Misra, Mairal, Goyal, Bojanowski, and Joulin]{a21}
Mathilde Caron, Ishan Misra, Julien Mairal, Priya Goyal, Piotr Bojanowski, and Armand Joulin.
\newblock Unsupervised learning of visual features by contrasting cluster assignments.
\newblock \emph{Proc. Adv. Neural Inf. Process. Syst.}, 33:\penalty0 9912--9924, 2020.

\bibitem[Chen et~al.(2021)Chen, Lu, Yu, Luo, Adeli, Wang, Lu, Yuille, and Zhou]{ax1}
Jieneng Chen, Yongyi Lu, Qihang Yu, Xiangde Luo, Ehsan Adeli, Yan Wang, Le Lu, Alan~L Yuille, and Yuyin Zhou.
\newblock Transunet: Transformers make strong encoders for medical image segmentation.
\newblock \emph{arXiv preprint arXiv:2102.04306}, 2021.

\bibitem[Chen et~al.(2018)Chen, Bentley, Mori, Misawa, Fujiwara, and Rueckert]{2222}
Liang Chen, Paul Bentley, Kensaku Mori, Kazunari Misawa, Michitaka Fujiwara, and Daniel Rueckert.
\newblock Drinet for medical image segmentation.
\newblock \emph{IEEE transactions on medical imaging}, 37\penalty0 (11):\penalty0 2453--2462, 2018.

\bibitem[Codella et~al.(2019)Codella, Rotemberg, Tschandl, Celebi, Dusza, Gutman, Helba, Kalloo, Liopyris, Marchetti, et~al.]{a15}
Noel Codella, Veronica Rotemberg, Philipp Tschandl, M~Emre Celebi, Stephen Dusza, David Gutman, Brian Helba, Aadi Kalloo, Konstantinos Liopyris, Michael Marchetti, et~al.
\newblock Skin lesion analysis toward melanoma detection 2018: A challenge hosted by the international skin imaging collaboration (isic).
\newblock \emph{arXiv preprint arXiv:1902.03368}, 2019.

\bibitem[Cong and Zhou(2023)]{a345}
Shuang Cong and Yang Zhou.
\newblock A review of convolutional neural network architectures and their optimizations.
\newblock \emph{Artif. Intell. Rev.}, 56\penalty0 (3):\penalty0 1905--1969, 2023.

\bibitem[Dosovitskiy et~al.(2020)Dosovitskiy, Beyer, Kolesnikov, Weissenborn, Zhai, Unterthiner, Dehghani, Minderer, Heigold, Gelly, et~al.]{a12}
Alexey Dosovitskiy, Lucas Beyer, Alexander Kolesnikov, Dirk Weissenborn, Xiaohua Zhai, Thomas Unterthiner, Mostafa Dehghani, Matthias Minderer, Georg Heigold, Sylvain Gelly, et~al.
\newblock An image is worth 16x16 words: Transformers for image recognition at scale.
\newblock \emph{arXiv preprint arXiv:2010.11929}, 2020.

\bibitem[Gehlot et~al.(2020)Gehlot, Gupta, and Gupta]{a346}
Shiv Gehlot, Anubha Gupta, and Ritu Gupta.
\newblock Ednfc-net: Convolutional neural network with nested feature concatenation for nuclei-instance segmentation.
\newblock In \emph{ICASSP}, pages 1389--1393. IEEE, 2020.

\bibitem[Gross(2002)]{a400}
Charles~G Gross.
\newblock Genealogy of the “grandmother cell”.
\newblock \emph{Neuroscientist.}, 8\penalty0 (5):\penalty0 512--518, 2002.

\bibitem[Gu et~al.(2018)Gu, Wang, Kuen, Ma, Shahroudy, Shuai, Liu, Wang, Wang, Cai, et~al.]{a344}
Jiuxiang Gu, Zhenhua Wang, Jason Kuen, Lianyang Ma, Amir Shahroudy, Bing Shuai, Ting Liu, Xingxing Wang, Gang Wang, Jianfei Cai, et~al.
\newblock Recent advances in convolutional neural networks.
\newblock \emph{Pattern Recognit.}, 77:\penalty0 354--377, 2018.

\bibitem[Gupta et~al.(2018)Gupta, Mallick, Sharma, Gupta, and Duggal]{a347}
Anubha Gupta, Pramit Mallick, Ojaswa Sharma, Ritu Gupta, and Rahul Duggal.
\newblock Pcseg: Color model driven probabilistic multiphase level set based tool for plasma cell segmentation in multiple myeloma.
\newblock \emph{PloS one}, 13\penalty0 (12):\penalty0 e0207908, 2018.

\bibitem[Gupta et~al.(2020)Gupta, Duggal, Gehlot, Gupta, Mangal, Kumar, Thakkar, and Satpathy]{a14}
Anubha Gupta, Rahul Duggal, Shiv Gehlot, Ritu Gupta, Anvit Mangal, Lalit Kumar, Nisarg Thakkar, and Devprakash Satpathy.
\newblock Gcti-sn: Geometry-inspired chemical and tissue invariant stain normalization of microscopic medical images.
\newblock \emph{Med. Image Anal.}, 65:\penalty0 101788, 2020.

\bibitem[Heidari et~al.(2023)Heidari, Kazerouni, Soltany, Azad, Aghdam, Cohen-Adad, and Merhof]{i18}
Moein Heidari, Amirhossein Kazerouni, Milad Soltany, Reza Azad, Ehsan~Khodapanah Aghdam, Julien Cohen-Adad, and Dorit Merhof.
\newblock Hiformer: Hierarchical multi-scale representations using transformers for medical image segmentation.
\newblock In \emph{WACV}, pages 6202--6212, 2023.

\bibitem[Hesamian et~al.(2019)Hesamian, Jia, He, and Kennedy]{a339}
Mohammad~Hesam Hesamian, Wenjing Jia, Xiangjian He, and Paul Kennedy.
\newblock Deep learning techniques for medical image segmentation: achievements and challenges.
\newblock \emph{J. Digital Imaging}, 32:\penalty0 582--596, 2019.

\bibitem[Huang et~al.(2020)Huang, Lin, Tong, Hu, Zhang, Iwamoto, Han, Chen, and Wu]{i4}
Huimin Huang, Lanfen Lin, Ruofeng Tong, Hongjie Hu, Qiaowei Zhang, Yutaro Iwamoto, Xianhua Han, Yen-Wei Chen, and Jian Wu.
\newblock Unet 3+: A full-scale connected unet for medical image segmentation.
\newblock In \emph{ICASSP}, pages 1055--1059. IEEE, 2020.

\bibitem[Huang et~al.(2022)Huang, Lin, Dou, Lin, Ying, Jia, Xu, Mei, Yang, Dong, et~al.]{a3}
Ruobing Huang, Mingrong Lin, Haoran Dou, Zehui Lin, Qilong Ying, Xiaohong Jia, Wenwen Xu, Zihan Mei, Xin Yang, Yijie Dong, et~al.
\newblock Boundary-rendering network for breast lesion segmentation in ultrasound images.
\newblock \emph{Med. Image Anal.}, 80:\penalty0 102478, 2022.

\bibitem[Isensee et~al.(2021)Isensee, Jaeger, Kohl, Petersen, and Maier-Hein]{a342}
Fabian Isensee, Paul~F Jaeger, Simon~AA Kohl, Jens Petersen, and Klaus~H Maier-Hein.
\newblock nnu-net: a self-configuring method for deep learning-based biomedical image segmentation.
\newblock \emph{Nat. Methods}, 18\penalty0 (2):\penalty0 203--211, 2021.

\bibitem[Li et~al.(2022)Li, Li, Rockwell, Farimani, and Lee]{i1}
Tianqin Li, Zijie Li, Harold Rockwell, Amir Farimani, and Tai~Sing Lee.
\newblock Prototype memory and attention mechanisms for few shot image generation.
\newblock In \emph{ICLR}, 2022.

\bibitem[Nam et~al.(2024)Nam, Syazwany, Kim, and Lee]{re3}
Ju-Hyeon Nam, Nur~Suriza Syazwany, Su~Jung Kim, and Sang-Chul Lee.
\newblock Modality-agnostic domain generalizable medical image segmentation by multi-frequency in multi-scale attention.
\newblock In \emph{Proceedings of the IEEE/CVF conference on computer vision and pattern recognition}, pages 11480--11491, 2024.

\bibitem[Oktay et~al.(2018)Oktay, Schlemper, Folgoc, Lee, Heinrich, Misawa, Mori, McDonagh, Hammerla, Kainz, et~al.]{a18}
Ozan Oktay, Jo Schlemper, Loic~Le Folgoc, Matthew Lee, Mattias Heinrich, Kazunari Misawa, Kensaku Mori, Steven McDonagh, Nils~Y Hammerla, Bernhard Kainz, et~al.
\newblock Attention u-net: Learning where to look for the pancreas.
\newblock \emph{arXiv preprint arXiv:1804.03999}, 2018.

\bibitem[Quiroga et~al.(2013)Quiroga, Fried, and Koch]{a401}
Rodrigo~Quian Quiroga, Itzhak Fried, and Christof Koch.
\newblock Brain cells for grandmother.
\newblock \emph{Sci. Am.}, 308\penalty0 (2):\penalty0 30--35, 2013.

\bibitem[Rahman and Marculescu(2023)]{i20}
Md~Mostafijur Rahman and Radu Marculescu.
\newblock Medical image segmentation via cascaded attention decoding.
\newblock In \emph{WACV}, pages 6222--6231, 2023.

\bibitem[Rahman et~al.(2024)Rahman, Munir, and Marculescu]{a20}
Md~Mostafijur Rahman, Mustafa Munir, and Radu Marculescu.
\newblock Emcad: Efficient multi-scale convolutional attention decoding for medical image segmentation.
\newblock In \emph{WACV}, pages 11769--11779, 2024.

\bibitem[Ramesh et~al.(2021)Ramesh, Kumar, Swapna, Datta, and Rajest]{1111}
KKD Ramesh, G~Kiran Kumar, K Swapna, Debabrata Datta, and S~Suman Rajest.
\newblock A review of medical image segmentation algorithms.
\newblock \emph{EAI Endorsed Transactions on Pervasive Health \& Technology}, 7\penalty0 (27), 2021.

\bibitem[Ronneberger et~al.(2015)Ronneberger, Fischer, and Brox]{i2}
Olaf Ronneberger, Philipp Fischer, and Thomas Brox.
\newblock U-net: Convolutional networks for biomedical image segmentation.
\newblock In \emph{MICCAI}, pages 234--241. Springer, 2015.

\bibitem[Santoro et~al.(2016)Santoro, Bartunov, Botvinick, Wierstra, and Lillicrap]{i22}
Adam Santoro, Sergey Bartunov, Matthew Botvinick, Daan Wierstra, and Timothy Lillicrap.
\newblock Meta-learning with memory-augmented neural networks.
\newblock In \emph{ICML}, pages 1842--1850. PMLR, 2016.

\bibitem[Shors(2008)]{a402}
Tracey~J Shors.
\newblock From stem cells to grandmother cells: how neurogenesis relates to learning and memory.
\newblock \emph{Cell Stem Cell}, 3\penalty0 (3):\penalty0 253--258, 2008.

\bibitem[Tang et~al.(2018{\natexlab{a}})Tang, Lee, Li, Zhang, Xu, Liu, Teo, and Jiang]{a7}
Shiming Tang, Tai~Sing Lee, Ming Li, Yimeng Zhang, Yue Xu, Fang Liu, Benjamin Teo, and Hongfei Jiang.
\newblock Complex pattern selectivity in macaque primary visual cortex revealed by large-scale two-photon imaging.
\newblock \emph{Curr. Biol.}, 28\penalty0 (1):\penalty0 38--48, 2018{\natexlab{a}}.

\bibitem[Tang et~al.(2018{\natexlab{b}})Tang, Zhang, Li, Li, Liu, Jiang, and Lee]{a8}
Shiming Tang, Yimeng Zhang, Zhihao Li, Ming Li, Fang Liu, Hongfei Jiang, and Tai~Sing Lee.
\newblock Large-scale two-photon imaging revealed super-sparse population codes in the v1 superficial layer of awake monkeys.
\newblock \emph{ELIFE}, 7:\penalty0 e33370, 2018{\natexlab{b}}.

\bibitem[Tu et~al.(2022)Tu, Talebi, Zhang, Yang, Milanfar, Bovik, and Li]{a341}
Zhengzhong Tu, Hossein Talebi, Han Zhang, Feng Yang, Peyman Milanfar, Alan Bovik, and Yinxiao Li.
\newblock Maxvit: Multi-axis vision transformer.
\newblock In \emph{ECCV}, pages 459--479. Springer, 2022.

\bibitem[Valanarasu et~al.(2021)Valanarasu, Oza, Hacihaliloglu, and Patel]{i9}
Jeya Maria~Jose Valanarasu, Poojan Oza, Ilker Hacihaliloglu, and Vishal~M Patel.
\newblock Medical transformer: Gated axial-attention for medical image segmentation.
\newblock In \emph{MICCAI}, pages 36--46. Springer, 2021.

\bibitem[Wang et~al.(2022{\natexlab{a}})Wang, Cao, Wang, and Zaiane]{i13}
Haonan Wang, Peng Cao, Jiaqi Wang, and Osmar~R Zaiane.
\newblock Uctransnet: rethinking the skip connections in u-net from a channel-wise perspective with transformer.
\newblock In \emph{AAAI}, pages 2441--2449, 2022{\natexlab{a}}.

\bibitem[Wang et~al.(2022{\natexlab{b}})Wang, Xie, Lin, Iwamoto, Han, Chen, and Tong]{i12}
Hongyi Wang, Shiao Xie, Lanfen Lin, Yutaro Iwamoto, Xian-Hua Han, Yen-Wei Chen, and Ruofeng Tong.
\newblock Mixed transformer u-net for medical image segmentation.
\newblock In \emph{ICASSP}, pages 2390--2394. IEEE, 2022{\natexlab{b}}.

\bibitem[Wang et~al.(2024)Wang, Lin, Wu, Yu, Cheng, and Yan]{i10}
Zhehao Wang, Xian Lin, Nannan Wu, Li Yu, Kwang-Ting Cheng, and Zengqiang Yan.
\newblock Dtmformer: Dynamic token merging for boosting transformer-based medical image segmentation.
\newblock In \emph{AAAI}, pages 5814--5822, 2024.

\bibitem[Woo et~al.(2018)Woo, Park, Lee, and Kweon]{i24}
Sanghyun Woo, Jongchan Park, Joon-Young Lee, and In~So Kweon.
\newblock Cbam: Convolutional block attention module.
\newblock In \emph{ECCV}, pages 3--19, 2018.

\bibitem[Xiao et~al.(2023)Xiao, Li, Liu, Zhu, and Zhang]{a340}
Hanguang Xiao, Li Li, Qiyuan Liu, Xiuhong Zhu, and Qihang Zhang.
\newblock Transformers in medical image segmentation: A review.
\newblock \emph{Biomed. Signal Process. Control}, 84:\penalty0 104791, 2023.

\bibitem[Xu et~al.(2023)Xu, Wang, Li, and Du]{a348}
Lianghui Xu, Liejun Wang, Yongming Li, and Anyu Du.
\newblock Big model and small model: Remote modeling and local information extraction module for medical image segmentation.
\newblock \emph{Appl. Soft Comput.}, 136:\penalty0 110128, 2023.

\bibitem[Xue et~al.(2021)Xue, Zhu, Fu, Hu, Li, Zhang, and Heng]{a2}
Cheng Xue, Lei Zhu, Huazhu Fu, Xiaowei Hu, Xiaomeng Li, Hai Zhang, and Pheng-Ann Heng.
\newblock Global guidance network for breast lesion segmentation in ultrasound images.
\newblock \emph{Med. Image Anal.}, 70:\penalty0 101989, 2021.

\bibitem[Yang and Chan(2018)]{i23}
Tianyu Yang and Antoni~B Chan.
\newblock Learning dynamic memory networks for object tracking.
\newblock In \emph{ECCV}, pages 152--167, 2018.

\bibitem[Yang and Farsiu(2023)]{re5}
Ziyun Yang and Sina Farsiu.
\newblock Directional connectivity-based segmentation of medical images.
\newblock In \emph{Proceedings of the IEEE/CVF conference on computer vision and pattern recognition}, pages 11525--11535, 2023.

\bibitem[Yao et~al.(2022)Yao, Hu, Li, Zhai, and Zhang]{i17}
Chang Yao, Menghan Hu, Qingli Li, Guangtao Zhai, and Xiao-Ping Zhang.
\newblock Transclaw u-net: claw u-net with transformers for medical image segmentation.
\newblock In \emph{ICASSP}, pages 280--284. IEEE, 2022.

\bibitem[Zhang et~al.(2021)Zhang, Liu, and Hu]{a343}
Yundong Zhang, Huiye Liu, and Qiang Hu.
\newblock Transfuse: Fusing transformers and cnns for medical image segmentation.
\newblock In \emph{MICCAI}, pages 14--24. Springer, 2021.

\bibitem[Zhou et~al.(2018)Zhou, Rahman~Siddiquee, Tajbakhsh, and Liang]{i3}
Zongwei Zhou, Md~Mahfuzur Rahman~Siddiquee, Nima Tajbakhsh, and Jianming Liang.
\newblock Unet++: A nested u-net architecture for medical image segmentation.
\newblock In \emph{Proc. Deep Learn. Med. Image Anal. Multimodal Learn. Clin. Decis Support.}, pages 3--11. Springer, 2018.

\bibitem[Zhu et~al.(2024)Zhu, Chen, Qiu, Farazi, Sotiras, Razi, and Wang]{re4}
Wenhui Zhu, Xiwen Chen, Peijie Qiu, Mohammad Farazi, Aristeidis Sotiras, Abolfazl Razi, and Yalin Wang.
\newblock Selfreg-unet: Self-regularized unet for medical image segmentation.
\newblock In \emph{International Conference on Medical Image Computing and Computer-Assisted Intervention}, pages 601--611. Springer, 2024.

\bibitem[Zhu et~al.(2023)Zhu, Cheng, Cui, Zhu, Ying, and Liang]{a1}
Yanjie Zhu, Jing Cheng, Zhuo-Xu Cui, Qingyong Zhu, Leslie Ying, and Dong Liang.
\newblock Physics-driven deep learning methods for fast quantitative magnetic resonance imaging: Performance improvements through integration with deep neural networks.
\newblock \emph{IEEE Signal Process Mag.}, 40\penalty0 (2):\penalty0 116--128, 2023.

\end{thebibliography}
}
\newpage
\clearpage
\setcounter{page}{1}
\maketitlesupplementary
\appendix
\section{Setting about $K$ in W-LD Update Strategy}
To accurately locate the category features of the input image, we propose the W-LD update strategy to dynamically update the prototype memory bank. The following is an introduction to the $K$ parameter in the W-LD update strategy.
\begin{figure}[ht]
\centerline{\includegraphics[width=0.43\textwidth]{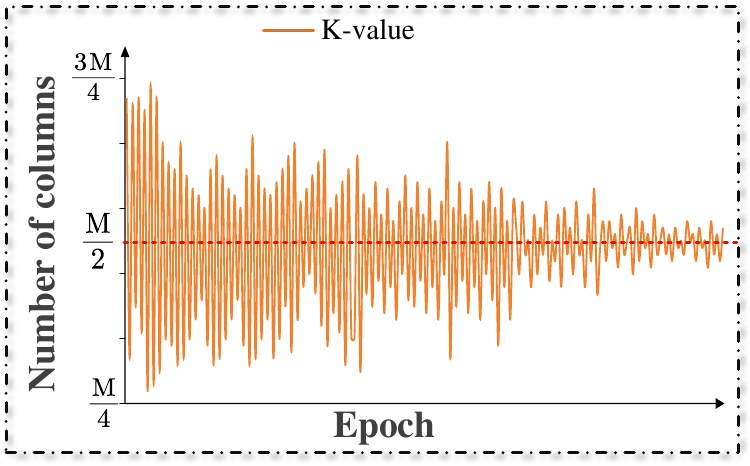}}
\caption{\textbf{Variation curve of $K$} on the Synapse dataset. $M$ is the feature dimension of similarity memory prior.}
\label{figure1}
\end{figure}
In W-LD update strategy, $K$ represents the update range of prototype memory bank. To prevent model training collapse, we add a restriction to $K$ ($\frac{M}{4}\leq  K\leq  \frac{3M}{4}$) to ensure effective updates to the prototype memory bank. It has been proven that our measure is effective, and there is no training collapse in all comparative experiments and ablation experiments. \Cref{figure1} show the curve of $K$ in the training process. It can be seen that with the increase of epoch, the model gradually converges, and $K$ is stable around $\frac{M}{2}$. In the initial stage of training, $K$ fluctuates violently, indicating that the model is learning useful patterns in data, and both large and small initial $K$ may have negative effects. Thus, we set the initial value of $K$ to $\frac{M}{2}$ in all experiments to cooperate with model training.

\section{Additional Visualization Results}
\label{B}
\begin{figure}[!t]
\centerline{\includegraphics[width=0.5\textwidth]{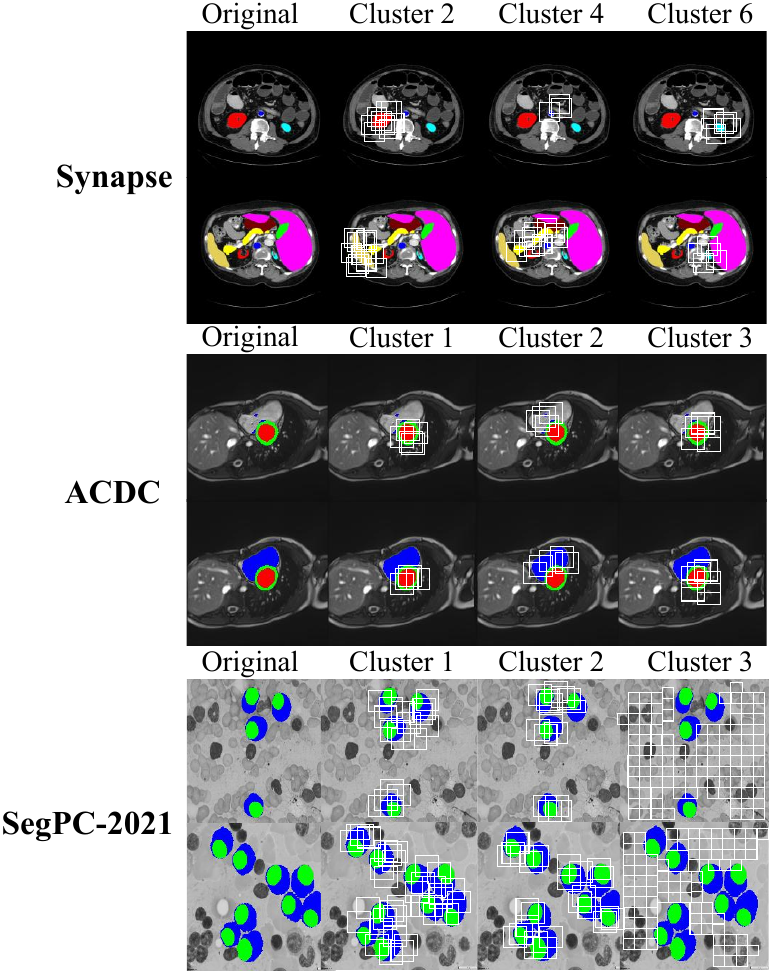}}
\caption{\textbf{Visualization of semantic clusters} (similarity memory priors) using DMW-LA on Synapse, ACDC, and SegPC-2021. We calculate the distribution of similarity memory prior in the prototype memory bank, and highlight their receptive fields with white boxes.}
\label{figure7}
\end{figure}

In the process of medical image segmentation, we assign a semantic cluster (similarity memory prior) to the segmentation target in each image to extract key categorical features. \Cref{figure7} shows the semantic cluster receptive field of DMW-LA on three datasets. We can observe that different clusters encode different semantic concepts. In the Synapse dataset, Cluster 2, Cluster 4 and Cluster 6 tend to focus on the region where the organs are located. In the ACDC dataset, Cluster 1 tends to focus on the left ventricle, while Cluster 2 focuses more on the right ventricle. In the SegPC-2021 dataset, Cluster 1 tends to focus on cytoplasm, while Cluster 3 focuses more on tissue fluid and other cells in the background. This distribution indicates that when processing images, each semantic cluster focuses on specific category features, thereby helping the model identify key structures in medical images.

\section{Evaluation Metrics}
\label{sec:3}
To better evaluate the performance of the network, we choose Dice score (DSC) and 95\% Hausdorff distance (HD95) as evaluation metrics, as shown in \Cref{eq3} and \Cref{eq4}.

\begin{equation}
\begin{aligned}
\label{eq3}
{DSC=\frac{(2*TP)}{(2*TP+FP+FN)}}
\end{aligned}
\end{equation}

\begin{equation}
\begin{aligned}
\label{eq4}
HD(A,B) = \max\{&\max_{a\in A}\{\min_{b\in B}\{d(a,b)\}\},\\
&\max_{b\in B}\{\min_{a\in A}\{d(b,a)\}\}\}
\end{aligned}
\end{equation}

In the above formula, $TP$ represents foreground information is predicted as foreground, $FP$ represents background information is predicted as foreground, and $FN$ represents foreground information is predicted as background. The HD95 represents the ${95}^{th}$ percentile of distances between the boundaries of A and B. DSC is used to evaluate the degree of overlap, while HD95 is used to evaluate distance. The combination of them can comprehensively evaluate the performance of the model.

\end{document}